# A Survey of Naïve Bayes Machine Learning approach in Text Document Classification

Vidhya.K.A
Department of Computer Science
Pondicherry University
Pondicherry, India

G.Aghila
Department of Computer Science
Pondicherry University
Pondicherry, India

*Abstract*— Text Document classification aims in associating one or more predefined categories based on the likelihood suggested by the training set of labeled documents. Many machine learning algorithms play a vital role in training the system with predefined categories among which Naïve Bayes has some intriguing facts that it is simple, easy to implement and draws better accuracy in large datasets in spite of the naïve dependence.

The importance of Naïve Bayes Machine learning approach has felt hence the study has been taken up for text document classification and the statistical event models available. This survey the various feature selection methods has been discussed and compared along with the metrics related to text document classification.

*Keyword- Text Mining, Naïve Bayes; Event models, Metrics, probability distribution.*

## I. INTRODUCTION

Text Document Classification is a task of classifying a document into predefined categories based on the contents of the document. A document is represented by a piece of text expressed as phrases or words. The task of traditional text categorization methods is done by human experts. It usually needs a large amount of time to deal with the task of text categorization. In recent years, text categorization has become an important research topic in machine learning and information retrieval and e-mail spam filtering. It also has become an important research topic in text mining, which analyses and extracts useful information from texts. More Learning techniques has been in research for dealing with text categorization. The existing text classification methods can be classified into below six [11],[12],[13] categories:

(1) Based on Rocchio's method (Dumais, Platt, Heckerman, & Sahami, 1998; Hull, 1994; Joachims, 1998; Lam & Ho, 1998).
(2) Based on K-nearest neighbors (KNN) (Hull, 1994; Lam & Ho, 1998; Tan, 2005; Tan, 2006; Yang & Liu, 1999).
(3) Based on regression models (Yang, 1999; Yang & Liu, 1999).
(4) Based on Naıve Bayes and Bayesian nets (Dumais et al., 1998; Hull, 1994; Yang & Liu, 1999; Sahami, 1996).
(5) Based on decision trees (Fuhr & Buckley, 1991; Hull, 1994).
(6) Based on decision rules (Apte`, Damerau, & Weiss, 1994; Cohen & Singer, 1999).

Among the six types the survey aims in getting an intuitive understanding of Naïve Bayes approach in which the application of various Machine Learning Techniques to the text categorization problem like in the field of medicine, e-mail filtering, including rule learning for knowledge base systems has been explored. The survey is oriented towards the various probabilistic approach of Naïve Bayes Machine Learning algorithm for which the text categorization aims to classify the document with optimal accuracy.

Naïve Bayes Model works with the conditional probability which originates from well known statistical approach "Bayes Theorem", where as Naïve refers to "assumption" that all the attributes of the examples are independent of each other given the context of the category. Because of the independence assumption the parameters for each attribute can be learned separately and this greatly simplifies learning especially when the number of attributes is large[15]. In this context of text classification, the probability that a document d belongs to class c is calculated by the Bayes theorem as follows

$$P(c/d) = \frac{P(d/c)P(c)}{P(d)} \quad (1)$$

The estimation of P (d/c) is difficult since the number of possible vectors d is too high. This difficulty is overcome by using the naïve assumption that any two coordinates of the document is statistically independent. Using this assumption the most probable category 'c 'can be estimated.

The survey is organized in the following depicted way that section II for Survey work where the discussion on probabilistic event modes are done, Section III for data characteristics affecting the Naïve Bayes model, Section IV for the results of Naive Bayes text classification method and Section V for Conclusion.

## II. SURVEY WORK

Despite its popularity, there has been some confusion in the document classification community about the "Naive Bayes" classifier because there are two *different* generative model in common use, both of which make the Naive Bayes assumption. One model specifies that a document is represented by a vector of binary attributes indicating which words occur and do not occur in the document. The number of times a word occurs in a document is not captured. When calculating the probability of a document, one multiplies the probability of all the attribute values, including the probability of non-occurrence for words that do not occur in the document. Here the document is considered to be the event," and the absence or presence of words to be attributes of the event. This describes the two models Multi-variate Bernoulli event model and Multinomial model as follows:

*A. Event Models For Naïve Bayes*

*Multi-variate Bernoulli Model:*
In the multi-variate Bernoulli Model a document is a binary vector over the space of words. Given a vocabulary *V*, each dimension of the space *t*, $t \in \{1,...,|V|\}$, corresponds to word $w_t$ from the vocabulary. Dimension 't' of the vector for document $d_i$ is written as $B_{it}$, and is either 0 or 1, indicating whether word $w_t$ occurs at least once in the document[6]. In such a document representation, the Naive Bayes assumption is made such that the probability of each word occurring in a document is independent of the occurrence of other words in a document [8]. Then, the probability of a document given its class from Equation 2 is simply the product of the probability of the attribute values over all word attributes:

$$P(d_i|c_j;\theta) = \prod_{t=1}^{|V|}(B_{it}P(w_t|c_j;\theta) +$$

$$(1-B_{it})(1-P(w_t|c_j;\theta))) \quad (1)$$

*Word Probability Estimate:*

$$\theta_{wt|cj} = P(w_t|c_j;\theta_j) = \frac{1+\sum_{i=1}^{|D|}B_{it}P(cj|di)}{2+\sum_{i=1}^{|D|}P(cj|di)} \quad (2)$$

*Maximumlikehood Estimate:*

$$P(c_j|\theta) = \frac{1+\sum_{i=1}^{|D|}P(cj|di)}{|D|} \quad (3)$$

*Working mode:*
This model does not capture the number of times each word occurs, and that it explicitly includes the non-occurrence probability of words that do not appear in the document.
To summarize, the definition of Naive Bayes learning algorithm is precisely given by describing the parameters that must be estimated, and how we may estimate them. When the n input attributes Xi each take on J possible discrete values, and Y is a discrete variable [10] taking on K possible values, then the learning task is to estimate two sets of parameters.
Estimation is done for these parameters using either maximum likelihood estimates (3) based on calculating the relative frequencies of the different events in the data or using Bayesian MAP estimates that is observed data with prior distributions over the values of these parameters.

*Multinomial model:*
In the multinomial model [10], a document is an ordered sequence of word events, drawn from the same vocabulary *V*. The assumption is made that the lengths of documents are independent of class. There again make a similar Naive Bayes assumption: that the probability of each word event in a document is independent of the word's context and position in the document. Thus, each document $d_i$ is drawn from a multinomial distribution of words with as many independent trials as the length of $d_i$. This yields the familiar "bag of words" representation for documents. Define $N_{it}$ to be the count of the number of times word $w_t$ occurs in document $d_i$. Then, the probability of a document given its class from Equation 5 is simply the multinomial distribution:

$$P(d_i|c_j;\theta) = P(|d_i|)|d_i|! \prod_{t=1}^{|V|} \frac{P(w_t|c_j;\theta)N_{it}}{N_{it}!} \quad (4)$$

*Word Probability Estimate:*

$$\theta_{wt|cj} = P(w_t|c_j;\theta_j) = \frac{1+\sum_{i=1}^{|D|}NitP(cj|di)}{|V|+\sum_{s=1}^{|V|}\sum_{i=1}^{|D|}Nis\,P(cj|di)} \quad (5)$$

*Maximumlikehood Estimate:*

$$P(c_j|d_i;\theta) = \frac{P(c_j|\theta)P(d_i|c_i;\theta_j)}{P(d_i|\theta)} \quad (6)$$

*Working Mode:*

In contrast to the multi-variate Bernoulli event model, the multinomial model captures word frequency information in documents. In case of continuous inputs Xi, we can of course continue to use equations (4) and (5) as the basis for designing a Naive Bayes classifier. However, when the Xi are continuous we must choose some other way to represent the distributions P(XijY). One common approach is to assume that for each possible discrete value yk of Y, the distribution of each continuous Xi is Gaussian, and is defined by a mean and standard deviation specific to Xi and yk. In order to train such a Naïve Bayes classifier the mean and standard deviation of each of these Gaussians should be estimated.

*Logistic Regression:*
Logistic Regression is an approach to learning functions of the form f: X! Y, or P (Yj|X) in the case where Y is discrete-valued, and X = hX1::Xni is any vector containing discrete or continuous variables. In this section the case where Y is a boolean variable is considered, in order to simplify notation. In the final subsection we extend our treatment to the case where Y takes on any finite number of discrete values. Logistic Regression [4] assumes a parametric form for the distribution P (Yj|X), then directly estimates its parameters from the training data. The parametric model assumed by Logistic Regression in the case where Y is boolean is:

$$P(Y=1|X) = \frac{1}{1+\exp(w_0 + \sum_{i=1}^{n}w_iX_i)} \quad (7)$$

$$P(Y=1|X) = \frac{\exp(w_0 + \sum_{i=1}^{n}w_iX_i)}{1+\exp(w_0 + \sum_{i=1}^{n}w_iX_i)} \quad (8)$$

One highly convenient property of this form for P(Y|X) is that is leads to a simple linear expression for classification. To

classify any given X such as to assign the value $y_k$ that maximizes $P(Y=y_k|X)$. [4]

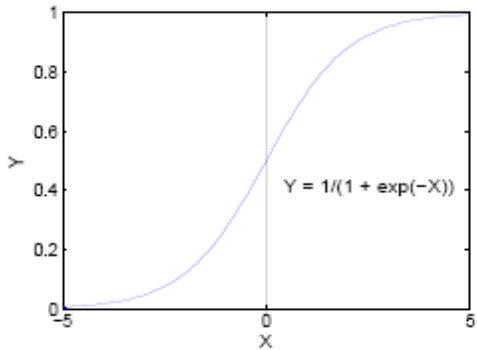

*Figure1. Logistic Regression Method.*

In Logistic Regression, $P(Y|X)$ is assumed to follow this form and take the natural log of both sides having a linear classification rule that assigns label Y=0 if it satisfies

$$0 < w_0 + \sum_{i=1}^{n} w_i X_i \qquad (9)$$

*B. Naïve Bayes with Active learning:*

Boosting is an iterative machine learning procedure [9] that successively classifies a weighted version of the instance, and then re-weights the instance dependent on how successful the classification was. Its purpose is to find a highly accurate classification rule by combining many weak or base classifiers, many of which may be only moderately accurate. The boosting method for Naïve Bayes determines the most appropriate class of the instance based on its current probability terms table[11]. There are various boosting techniques for learning to overcome the noise in the data in which Naïve Bayes machine learning methodology along with the active learning method improves the classification accuracy.

*Working Mode:*

A set of 15 data sets from the UCI machine learning repository are considered for the [9] experiments. A pre-discretized step based on entropy algorithm was applied to data sets that include continuous attributes, [9] which converts continuous attributes into nominal attributes for boosting. In data sets with missing value, the consideration the most frequent attribute value as a good candidate from which the Naïve Bayes learning technique had a better classification accuracy.

*C. Naïve Bayes Classification and PET*

Different from Normal Decision Tree more properly pruning reduces the performances of Probability Estimation Tree (PET) in order to get good probability estimation where large trees are required however it doesn't support the model transparency [3], [14]. Given a PET *T* which is learnt from D, according to the Baseyian theorem a data element x= <x1, x2,…xm> can be classified

$$P(C_k|x,T) \propto P(x|C_k,T)P(C_k,T) \qquad (10)$$

The attributes are divided into two disjoint groups denoted by $x_T = \{x1,x2,,…..xm\}$ and $x_B = \{xm+1,…….xn\}$ respectively. $x_T$ be the vector of variables contained in the given tree T and the rest variables are contained in $x_B$ Under this independence assumption, the following equation is derived along with Bayes theorem,

$$P(x|C_k) = \frac{P(C_k|x_T,T)P(xT|T)}{P(C_k|x_T)} \prod_{j \in xB} P(x_T|C_k)$$

(11)

*Working Mode:*

Given a training dataset, a small-sized tree can be learnt using single PET in which classes are split evenly in the dataset considered. Bayesian Estimated PET (BPET) model generally performs better at the shallow depths than the PET Model.

*D. Naïve Bayes Classification and Maximum Entropy Model.*

To achieve a classification accuracy of English texts, Naïve Bayes [1], [2] method based on base noun phrase (BaseNP) identification along with the rising maximum entropy model is applied to the identification for best features in the document. Maximum entropy model is a quite mature statistical model, which adapts to evaluate the probability distribution of events. For BaseNP identification problem, a word may be viewed as an event and the context of this word mav be viewed as the environment of this event.

*Feature Selection Method:*

Firstly, use training corpus and user-defined feature templates to generate candidate features. Secondly, the feature selection algorithm computing feature gains is applied to select features. Finally, at the parameter estimation stage, the improved iterative scaling (IIS) algorithm is adopted.

*Working Mode:*

The experimental results show that this technique achieved precision and recall rates of roughly 93% for BaseNP identification and the classification accuracy is remarkably improved on this basis. It indicates that shallow parsing of high accuracy is very helpful to text classification.

III. DATA CHARACTERISTICS AFFECTING NAÏVE BAYES PERFORMANCE

Naïve Bayes works well for the data characteristics with certain deterministic or almost deterministic dependencies that is low entropy distribution, however the intriguing fact is that algorithm work well even when the independence assumption is violated [5]. To address the above issue Naïve Bayes optimality is checked with the zero-Bayes risk problem to demonstrate empirically that the entropy $P(x_i|0)$ is a better predictor of the Naïve Bayes error than the class-conditional mutual information between features. There are some data characteristics for which the Naïve Bayes works as follows,

Monte Carlo simulations, is used to show that Naïve Bayes works best in two cases: completely independent [5] features (as expected by the assumptions made) and functionally dependent features. Naive Bayes has its worst performance between these extremes.

*Zero-Bayes Risk Problem:*

For the above mentioned reason in order to prove Naive Bayes optimality (Domingos & Pazzani, 1997) for some problems classes that have a high degree of feature dependencies, such as disjunctive and conjunctive concepts are studied. The data characteristics are explored that make naive Bayes work well, for zero-Bayes-risk problems, [5] it has been proved the Naive Bayes optimality for any two-class concept with nominal features where only one example has class 0 (or class 1), thus generalizing the results for conjunctive and disjunctive concepts. Then, using Monte-Carlo simulation, the behaviour of Naïve Bayes for increasing prior probability was studied and compared.

*Working Mode:*

Naive Bayes classifier is optimal for any two class concept with nominal features that assigns class 0 to exactly one example and class 1 to the other examples, [5] with probability 1. Thus entropy of class-conditional marginal is a better indicator of Naïve Bayes error than the mutual information between the features. However, the variance of such prediction is quickly increasing with and is quite high when gets closer to 0.5.

## IV. EXPERIMENTAL METHODOLOGY- RESULTS WITH DATASETS

Naïve Bayes generally outperforms for large datasets in text classification problem in spite of the Naïve independence assumption but as of small data sets Naïve Bayes doesn't show promising results in accuracy or performance.[6] Even though Naïve Bayes technique achieves better accuracy, to fine tune the classification accuracy it's combined with the other machine learning technique like SVM, neural networks, decision trees which has been discussed above.

Basically Naïve Bayes work with the conditional probability derived from the idea of Bayes theorem which is modified according to the application of Naïve Bayes for text classification. To evaluate the text classifier system with the Naïve Bayes approach there are two metrics factor, precision, recall and F1-measure can be used to find the effectiveness of document classifier which is given by, tp (True Positive): The number of documents correctly classified to that class.

tn (True Negative): The number of documents correctly rejected from that class.

fp (False Positive): The number of documents incorrectly rejected from that class.

fn (False Negative): The number of documents incorrectly classified to that class.

$$P: \text{Precision} = tp/(tp+fp) \quad (12)$$
$$R: \text{Recall} = tp/(tp+fn) \quad (13)$$
$$F1 Measure = 2.(P.R)/(P+R) \quad (14)$$

The formulas for precision, recall and F-measure is given in (12), (13), (14). The performance of Naïve Bayes Machine learning technique when combined with the other method shows better performance.

The discussion about the Naïve Bayes performance with the micro F1-measure values for the multinomial methods available from paper [15] such that the variants of the classifiers significantly outperform the traditional multinomial Naive Bayes at least when the 20-Newsgroup is used. In the graph showing the microF1 values, SRF _ l at _of 0.2 achieves the best performance. RF _ u and SRF _ u also achieve better performance [15] than baseline performance and less so than the Rf _ l or SRF _ l, but trivial. It means that there is no significant difference between using the number of tokens and the number of unique terms.

The biggest difference between the microF1 and macroF1 is that the performance increase by the normalization over the baseline is much greater in the case of macroF1 (0.2238 for the baseline versus 0.5066 for RF-l). Since macroF1 values in the Reuters21578 collection tend to be dominated by a large number of small categories, which have a small number of training documents [15], From the above survey of this paper it is understood that the proposed normalization methods are quite effective, particularly in the categories where the number of positive training documents is small where the traditional Naïve Bayes Technique fails, the author have done subsequent experiments and found the method is quite effective.

For Text categorization there are various benchmark datasets available like Reuters21578, Cora, WebKB and 20Newsgroup. Reuters21578 and 20Newsgroup datasets are designed with either set of long or short document. There are predefined categories where the hierarchy structure for each category is specified. The dataset WebKB is generally preferred for spam mail classification simulation. However the results of the Naïve Bayes along with the other hybrid methods for text document classification with these datasets and feature selection technique is depicted in the following Table1. Performance of Naïve Bayes when combined with other methods,

TABLE I. PERFORMANCE OF NAÏVE BAYES WHEN COMBINED WITH OTHER TECHNIQUES

| *Text Document Classification and Naïve Bayes Machine Learning Approach* | | | |
|---|---|---|---|
| **Naïve Bayes Model (Method)** | **Feature Selection Techniques** | **DataSets Used** | **Accuracy/Performance** |
| Naive Bayes Model with Noun Phrase approach | User defined Feature selection template | The training material comes from four sections (sections 15-18) of the Wall Street Journal (WSJ) part of the Penn Treebank-II corpus, including 400 English texts composed of 211727 words. Theother three sections (sections 20-22) are separately as the test material. | 93.7% |
| Naïve Bayes with Probability estimation Tree | Small Size Data –No Feature Selection Required | Experiments on 9 UCI Datasets are conducted | On Average 87% |
| Naïve Bayes with Support Vector Machine | TF-IDF (Term Frequency and Inverse Term Frequency Method) | 20Newgroup and Prepared own Dataset for testing | Flat Ranking – 88.89% Flat Ranking with High Ranking Keyword – 90.00% |
| Naïve Bayes with Active Learning Boosting Method | Weightage Scheme | 15 Datasets from the UCI Machine Learning Repository | Achieved Higher Accuracy by 0.05% compared to Adaboost |
| Naïve Bayes with Generative/Discriminative Technique | Wavelet transformation Feature subset of Documents | Reuters21578, Cora, WebKB and 20Newsgroup Dataset | 92.5% on Average |
| Naïve Bayes for Learning Object Identification | Weightage Scheme, Normalized Statistics | Set of own data files | Good Learning Object Identification is achieved. |
| Naïve Bayes for E-mail Spam Filtering | Mutual Information Gain | Lingspam corpus and PUI corpus | Multivariate – 98.86% Accuracy ,Multinomial-98.06% |
| Naïve Bayes with Multivariate and Multinomial Distribution | New Feature Weightage scheme was proposed and tested | Reuters21578 and 20Newsgroup | F1-Measure is compared for various weightage scheme. Poisson -0.5066 Multinomial-0.2238 |

CONCLUSION

Text Document Classification has been in research for decade in which various researchers has experimented with available machine learning techniques in which each method has been aimed to improve the classification accuracy; Among which Naïve Bayes works well in large datasets even with the simple learning algorithm had been a great inspirations in doing this survey. From the survey the inference made is that the Naïve Bayes technique performs better and yields higher classification accuracy when combined with the other techniques. The other inference is that Multinomial Naïve Bayes event model is more suitable when the dataset is large when compared to the Multi-variate Bernoulli Naïve Bayes Model.

REFERENCES

[1] "Yu-Chuan Chang, Shyi-Ming Chen, Churn-Jung Liau. "Multilabel text categorization based on a new linear classifier learning method and a category-sensitive refinement method". Expert Systems with Applications 34 (2008).

[2] Lin Lv,Yu-Shu Liu, "Research and realization of naïve bayes english text classification method based on base noun phrase identification," School ofComputer Science and Technology, Beijing Institute ofTechnology, Beiang 100081, China

[3] Zengchang Qin "Naive Bayes Classification Given Probability estimation Trees" Proceedings of the 5th International Conference on Machine Learning and Applications (ICMLA'06).

[4] www.cs.cmu.edu/_tom/mlbook.html - Chapter1

[5] Irina Rish, Joseph Hellerstein , Jayram Thathachar "An analysis of data characteristics that affect naive Bayes performance" 2001.

[6] Akinori Fujino, Naronori Ueda and Kazumi saito "Semisupervised Learning for a Hybrid Generative/Discriminative Classifier Based on the maximum Entropy Principle" IEEE Transactions on Pattern Analysis and Machine Intelligence. March 2008.

[7] Dino Isa, Lam Hong Kee, V.P. kallimani and R.Rajkumar "Text Document Pre-processing with Bayes formula for classification using SVM" IEEE Transactions on Knowledge and Data Engineering -2008.

[8] Karl-Michael Schneider,"A Comparison of event Models for Naïve Bayes Anti-spam E-mail Filtering" 2003.

[9] Li-mm wang', Sen-mia yuan', Ling Liz, Hai-Jun Liz "Boosting Navie Bayes by Active Learning"

[10] A McCallum, K.Nigam. A comparison of event models for naïve Bayes text classification. AAAA-98 workshop on Learning for text Categorization, 2004.

[11] Tom M.Mitchell, "Machine Learning," Carnegie Mellon University, McGraw-Hill Book Co, 1997.

[12] YI-Hsing Chang, Hsiu-Yi Huang, "An automatic document classification Based on Naïve Bayes Classifier and Ontology".Proceedings of the seventh International conference on Machine Learning and Cybernetics,2008.

[13] Vishal Gupta , Gurpreet S. Lehal "A Survey of Text Mining Techniques and Applications" Journal of Emerging Technologies in Web Intelligence, VOL 1, No.1, August 2009.

[14] "Tzung-Shi Chen and Shih-Chun Hsu"Mining frequent tree-like patterns in large datasets "Data & Knowledge Engineering August 2006.

[15] Sang-Bum kim, Kyong-soo Han, Hae-Chang Rim, Sung Hyon Myaeng "Some Effective techniques for Naïve Bayes Text Classification" IEEE Transactions on Knowledge and Data Engineering -2006.